\documentclass[letterpaper, 10 pt, journal, twoside]{IEEEtran} 

\IEEEoverridecommandlockouts                              

\usepackage{amsmath} 
\usepackage{amssymb}  
\usepackage[colorinlistoftodos]{todonotes}
\usepackage{hyperref}
\usepackage{multirow}
\usepackage{booktabs}
\usepackage[colorinlistoftodos]{todonotes}
\usepackage{siunitx}
\usepackage{url}

\newcommand{\etal}{\textit{et al}. }
\newcommand{\ie}{\textit{i}.\textit{e}., }

\usepackage[normalem]{ulem}

\usepackage{accents}
\newcommand{\ubar}[1]{\underaccent{\bar}{#1}}

\newcommand{\vth}{\mbox{\boldmath $\theta$}}

\newcommand{\vlambda}{\mbox{\boldmath $\lambda$}}

\newcommand{\vnu}{\mbox{\boldmath $\nu$}}

\newcommand{\vom}{\mbox{\boldmath $\omega$}}

\newcommand{\vf}{\mathbf f}
\newcommand{\vg}{\mathbf g}

\newcommand{\vn}{\mathbf n}

\newcommand{\vp}{\mathbf p}
\newcommand{\vq}{\mathbf q}
\newcommand{\vr}{\mathbf r}
\newcommand{\vs}{\mathbf s}

\newcommand{\vu}{\mathbf u}
\newcommand{\vv}{\mathbf v}

\newcommand{\vx}{\mathbf x}

\newcommand{\vA}{\mathbf A}
\newcommand{\vB}{\mathbf B}
\newcommand{\vC}{\mathbf C}
\newcommand{\vD}{\mathbf D}

\newcommand{\vI}{\mathbf I}

\newcommand{\vQ}{\mathbf Q}
\newcommand{\vR}{\mathbf R}

\newcommand{\vT}{\mathbf T}

\title{
Frequency-Aware Model Predictive Control
}

\author{Ruben Grandia$^1$, Farbod Farshidian$^1$, Alexey Dosovitskiy$^2$, Ren\'{e} Ranftl$^2$, Marco Hutter$^1$
\thanks{Manuscript received: September, 10, 2018; Revised December, 10, 2018; Accepted January, 11, 2019.}
\thanks{This paper was recommended for publication by Editor Nikos Tsagarakis upon evaluation of the Associate Editor and Reviewers' comments. 
This research was supported by Intel Network on Intelligent
Systems, the Swiss National Science Foundation through the National Centre of Competence in Research Robotics (NCCR Robotics), the European Union’s Horizon 2020 research and innovation programme under grant agreement No 780883. This work has been conducted as part of ANYmal Research, a community to advance legged robotics.}
\thanks{ $^1$ First, second, and last authors are with Robotic Systems Lab, ETH Zurich, Switzerland
        {\tt\footnotesize rgrandia@ethz.ch}.
        $^2$ Third and fourth authors are with Intel Labs, Munich, Germany.}%
\thanks{Digital Object Identifier (DOI): see top of this page.}
}

\begin{document}

\maketitle

\begin{abstract}
Transferring solutions found by trajectory optimization to robotic hardware remains a challenging task.
When the optimization fully exploits the provided model to perform dynamic tasks, the presence of unmodeled dynamics renders the motion infeasible on the real system.
Model errors can be a result of model simplifications, but also naturally arise when deploying the robot in unstructured and nondeterministic environments.
Predominantly, compliant contacts and actuator dynamics lead to bandwidth limitations. 
While classical control methods provide tools to synthesize controllers that are robust to a class of model errors, such a notion is missing in modern trajectory optimization, which is solved in the time domain. 
We propose frequency-shaped cost functions to achieve robust solutions in the context of optimal control for legged robots. 
Through simulation and hardware experiments we show that motion plans can be made compatible with bandwidth limits set by actuators and contact dynamics.
The smoothness of the model predictive solutions can be continuously tuned without compromising the feasibility of the problem.
Experiments with the quadrupedal robot ANYmal, which is driven by highly-compliant series elastic actuators, showed significantly improved tracking performance of the planned motion, torque, and force trajectories and enabled the machine to walk robustly on terrain with unmodeled compliance.
\end{abstract}


\section{Introduction}
\begin{figure}[!t]
    \centering
    \includegraphics[width=0.7\columnwidth]{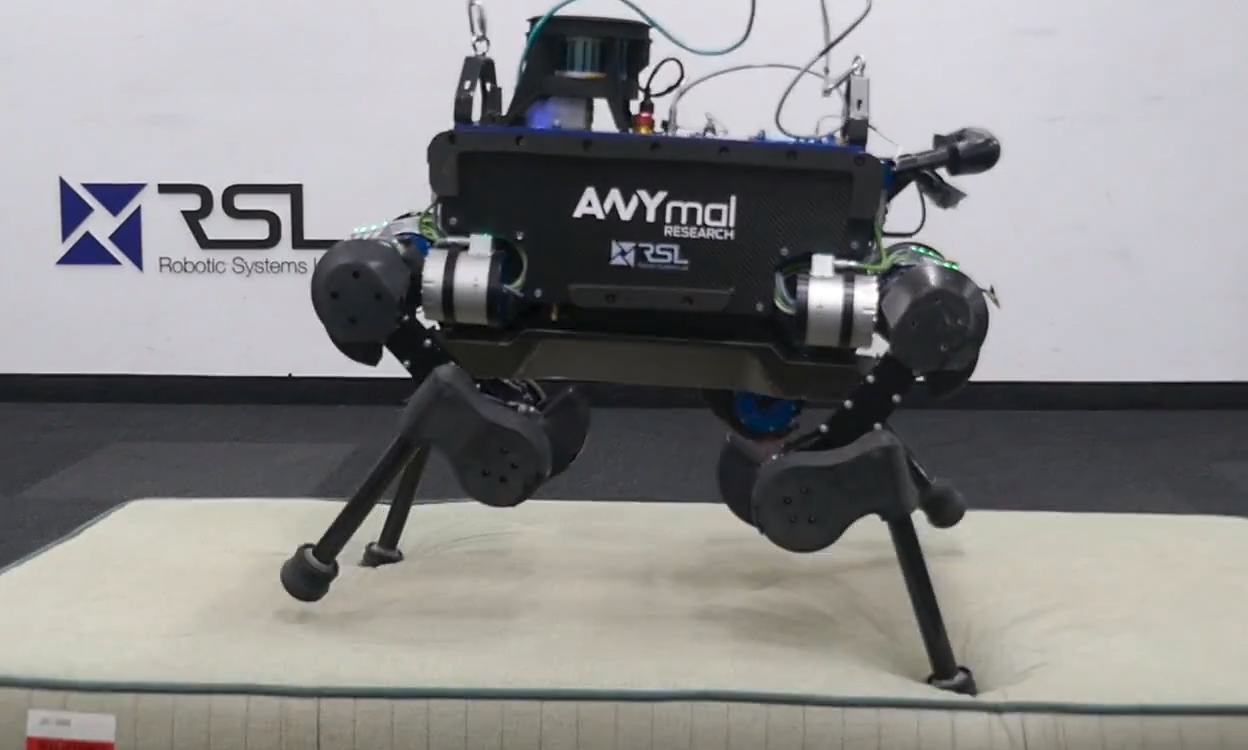}
    \vspace{-2mm}
    \caption{The quadruped robot ANYmal \cite{hutter2016anymal} trotting in place on non-rigid terrain. This experimental setup is used to test the controller's robustness against unmodeled contact dynamics.}
    \label{fig:anymal}
    \vspace{-5mm}
\end{figure}

\IEEEPARstart{T}{rajectory} optimization based on the full dynamics of a robotic system provides a flexible tool to generate complex motion plans. 
It enables the system to exploit the dynamic capabilities of the robot to achieve a task.
State-of-the-art approaches are able to rapidly find solutions while incorporating increasingly complex model descriptions, which allows using trajectory optimization in a Model Predictive Control (MPC) fashion. 
However, relying on the specific structure of the model makes implementation of the synthesized motion plans prone to modeling errors.
Executing motion plans on hardware has therefore proven to be nontrivial and often requires manual, task-dependent tuning of cost functions and constraints to achieve feasible motions. 
A major source of modeling error is the treatment of actuators as perfect torque sources. 
Any real system is subject to bandwidth limits and as such is not an ideal torque source.
A similar modeling error occurs when assuming a rigid contact with the ground. 
The rigid contact essentially provides the optimizer with infinite bandwidth control over the contact forces.
This assumption generally does not hold during locomotion in outdoor environments or on compliant surfaces as shown in Fig.~\ref{fig:anymal}.
As a result, motion plans generated assuming idealized contact and actuator dynamics cannot be tracked by the hardware, leading to poor tracking performance or failure of the locomotion controller. 

In this paper, we extend MPC methods for legged locomotion to situations where the assumptions of rigid ground and perfect actuators are invalid. 
The selected baseline model describes the 6 degrees of freedom Center of Mass (CoM) dynamics and motion of each leg.
Simultaneous optimization of footstep location and contact interaction is achieved by having both contact forces and joint velocities as control inputs.
We address the issues of inherent bandwidth limits in real robots by adapting the cost function to be frequency-dependent, making it possible to penalize high frequencies in the motion plans.
The solver, therefore, does not have to reason about the exact details of terrain and actuator dynamics but will produce solutions that are achievable under the bandwidth limits. 
%
%
%
We show that motion plans generated with our frequency-aware trajectory optimization can be followed by the hardware more accurately than those generated with a standard baseline and enables locomotion on compliant terrain.

\subsection{Related work}
Local feedback stabilization around a planned motion is a well-known technique to mitigate modeling errors \cite{Takenaka2009,Englsberger2012,Zhou2016,Sygulla2017} and has led to successful soft ground walking for a bipedal robot~\cite{Hashimoto2012}.
However, especially for dynamic motions, performance can be increased by providing a high-quality feedforward term. 
The effort in this work to improve the feasibility of the feedforward term and state reference can be seen as complementary to local stabilization strategies.
Moreover, disturbances can be rejected by adjusting the motion plan through fast replanning, reducing the burden on the feedback controller.
%

In the case of a series elastic actuator, the dynamics can be approximated and added to the model. 
The optimization algorithm is in those cases able to exploit the properties of the specific actuator and adding spring-damper elements to the joints is even known to result in motions that resemble those found in nature~\cite{Schultz2010}.
For series elastic actuators, methods have been proposed to incorporate bandwidth, torque, and joint limits in a computationally efficient way~\cite{Nakanishi-RSS-12}\cite{Braun2013}\cite{schlossman2018exploiting}.
However, very often we do not have exact details of actuators and modeling them would lead to high engineering effort. 
Moreover, since parts of the underlying actuator dynamics have very different time constants, smaller timesteps are required.
This leads to slower update rates, preventing such models from use in MPC with complex systems.
Additionally, stability problems can arise when a limb with stiffly modeled actuators makes contact with the environment~\cite{Koenemann2015}.
We avoid such issues by not explicitly modeling the actuators, but by incorporating well-known bandwidth limitations up to which the perfect tracking assumption is valid.
%

While model parameters for actuators can be obtained from first principles or through repeated experiments, contact dynamics are considerably harder to model or predict. 
A combination of learning a terrain model and trajectory optimization has been proposed in~\cite{Chang2017}.
However, such methods have not yet reached real-time capabilities. 
In the context of contact invariant optimization, soft contact models~\cite{Neunert2017} or contact smoothing~\cite{mordatch2012discovery} are used inside trajectory optimization.
These models are selected and tuned for their numerical properties rather than physical accuracy.
The models need to be smooth since the highly coupled interaction between stiff contact dynamics and slow dynamics of the robot lead to poor convergence of the algorithms. 
In the worst case, this numeric model tuning can lead to highly undesired effects when the optimizer starts exploiting dynamic properties of the terrain which are entirely wrong.

Reasoning over higher order terms results in solutions with a higher degree of continuity, which improves performance on hardware considerably.
This can be achieved by selecting a smooth parameterization of the solution space as done
in spline based optimization~\cite{Kalakrishnan2010,bellicoso2018dynamic,Werner2017_IROS}, collocation methods~\cite{Hereid2016,Pardo-RSS-17}, or when using dynamic motion primitives~\cite{Werner2017}.
This, however, limits the motions that can be expressed and can often require a problem-specific, manual tuning procedure.
Alternatively, higher derivatives can be selected as the control inputs in MPC \cite{Kajita2003}.
In Sect.~\ref{sect:relatedmethods} we show that this formulation can be interpreted as a special case of the presented frequency weighting method.
As an alternative to higher order formulations or explicitly modeling actuator dynamics, we propose to encode bandwidth limits through the cost function.
A trivial way to do so is to put extra costs on input signals, but this results in slower response overall and goes against the desire to perform highly dynamic motions at the limits of the system.
Instead, we intend to explore a different approach and propose to use a frequency-dependent cost function \cite{gupta1980frequency}. 
We penalize control actions only in the high frequency range and combine this idea with a modern optimal control framework \cite{farshidian2017efficient,Farshidian2017MPC}.

Frequency-based approaches have been proposed with several applications in both the MPC and legged robotic communities.
In \cite{hours2015constrained} constraints on the output spectrum are formed in the frequency domain. 
However, the proposed window-based constraints approach requires previous and future decision variables, which negatively affects the Riccati-sparsity pattern exploited in optimal control methods.
In \cite{Hashimoto2015}, a Fourier transform has been used
to find a closed form solution for momentum compensation of the lower body with the upper body.
In contrast to our setting where the unmodeled dynamics are unknown, the forces that are to be compensated for are assumed to be known based on a full rigid body dynamics model.

\subsection{Contributions}
We introduce frequency-dependent cost functions integrated into modern MPC strategies for legged locomotion.
Through simulation experiments, we study the effect of such a cost function on the resulting solutions.
The proposed method provides the user with an intuitive way to achieve robustness against unmodeled phenomena like actuator bandwidth limits and non-rigid contact dynamics.
%
%
These findings were successfully validated in hardware experiments on different grounds.
Using frequency-shaped cost functions, we could improve the robustness of ANYmal while locomoting under substantial external disturbances coming from external pushes or unmodeled soft ground.

\section{Method}
First, we discuss uncertainty in the dynamics from a robustness point of view and motivate the particular choice of cost functions.
Afterward, the integration with a Sequential Linear Quadratic Model Predictive Control (SLQ-MPC) method \cite{Farshidian2017MPC} is presented.
This method, based on Differential Dynamic Programming, relies on a linear approximation of the dynamics and a quadratic approximation of the cost function around the latest trajectory.
For brevity of notation in the current section, and without loss of generality, we consider the following quadratic cost function without linear and mixed state-input costs,
\begin{equation}
 J = \frac{1}{2} \int_0^{\infty} { \left( \vx(t)\!^T\! \vQ \vx(t) +  \vu(t)\!^T \vR \vu(t) \right) \, dt},
    \label{eq:J_standard_timedomain}
\end{equation}
where $\vQ$ is the positive semi-definite state cost Hessian and $\vR$ is the positive definite input cost Hessian. 

\subsection{High frequency robustness}
\begin{figure}[tb]
    \centering
    \includegraphics[width=0.7\columnwidth]{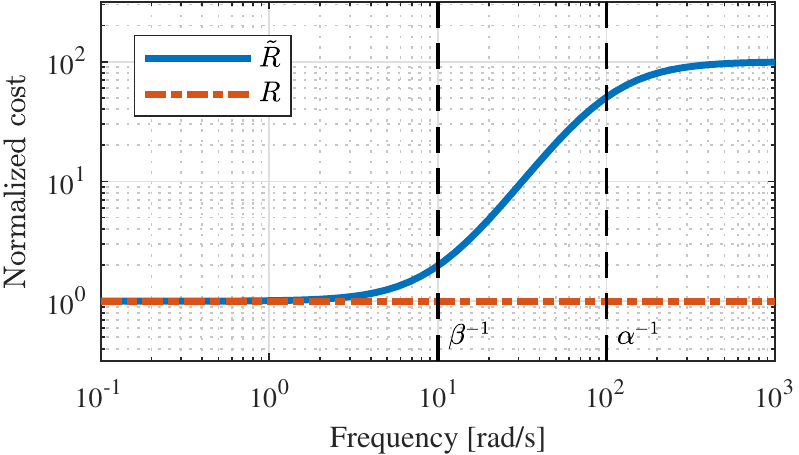}
    \vspace{-3mm}
    \caption{Example of the frequency-shaped cost function for $\alpha=0.01$, $\beta=0.1$, and the standard input costs in the frequency domain. Costs are normalized by $R$.}
    \label{fig:costfunctions}
    \vspace{-5mm}
\end{figure}

Consider a linear plant $G(j\omega)$ with unstructured multiplicative uncertainty model $L(j\omega)$,
\begin{equation}
\tilde{G}(j\omega) = [I + L(j\omega)] G(j\omega),
\end{equation}
\begin{equation}
\bar{\sigma}[L(j\omega)] < l_m( \omega ), \quad \forall \omega \geq 0,
\end{equation}
where $\bar{\sigma}$ is the maximum singular value of the disturbance model and $l_m( \omega )$ is a frequency-dependent upper bound.
The closed loop stability condition for these models is \cite{doyle1981multivariable},
\begin{equation}
l_m( \omega ) < \ubar{\sigma}[I + GK(j\omega)^{-1}],
\label{eq:robustStabilityCriteria}
\end{equation}
where $\ubar{\sigma}$ is the minimum singular value, and $GK(j\omega)$ is the transfer function of plant and controller together.
To be robust against large uncertainties at high frequencies, according to \eqref{eq:robustStabilityCriteria}, the loop gain, $GK(j\omega)$, should be kept low.
Intuitively, penalizing inputs at the high frequencies reduces the feedback gain at those frequencies, which allows for larger uncertainty magnitude, $l_m( \omega )$.
We therefore propose to use the following frequency-dependent input weighting
\begin{equation}
    \tilde{R}(\omega) = \left|\frac{1 + \beta j\omega}{1 + \alpha j\omega}\right|^2 R, \qquad \text{with}\quad\beta>\alpha,
    \label{eq:FreqShapedCost}
\end{equation}
where $R$ is the original input cost, and $-\beta^{-1}$ and $-\alpha^{-1}$ are the zero and pole of the loopshaping transfer function.
A visualization of such cost function is provided in Fig.~\ref{fig:costfunctions}.

Indeed, for Single-Input Single-Output (SISO) systems Anderson \etal \cite{anderson1985use} established that the open loop gain at high frequency under the frequency-shaped cost function \eqref{eq:FreqShapedCost} is reduced, \ie $|GK_{\tilde{R}}(j\omega)| < | GK_{R}(j\omega)|$ for large $\omega$, where $K_{\tilde{R}}$ and $K_R$ are the Linear Quadratic Regulator gains obtained under the frequency-shaped cost and baseline cost respectively.
According to \eqref{eq:robustStabilityCriteria}, the resulting increase in $\ubar{\sigma}[I + GK(j\omega)^{-1}]$ permits a higher model uncertainty in the stopband.
Unfortunately, to the best of our knowledge, a robustness proof for Multiple-Input Multiple-Output systems is not available.
Despite that, the intuition that penalizing high frequency input increases compatibility with actuator bandwidth limits remains.
In this paper, we aim to empirically validate the effect of using such a cost function.

\subsection{Frequency-shaped cost functions}
MPC plans over a receding horizon.
The cost function in \eqref{eq:FreqShapedCost} therefore needs to be expressed in the time domain as well.
This can be achieved by a state augmentation as described in~\cite{gupta1980frequency}.
The standard quadratic cost function for the time domain  \eqref{eq:J_standard_timedomain} can be converted to the frequency domain \eqref{eq:J_standard_frequencydomain} according to Parseval's theorem:
\begin{align}
    J &= \frac{1}{2\pi} \int_{-\infty}^{\infty} \left( \hat{\vx}(\omega)^H \vQ \, \hat{\vx}(\omega) 
    + \hat{\vu}(\omega)^H \vR \, \hat{\vu}(\omega) \right) \,d\omega ,
    \label{eq:J_standard_frequencydomain}
\end{align}
where $\hat{\vx}(\omega)$, and $\hat{\vu}(\omega)$ are the Fourier transform of $\vx(t)$ and $\vu(t)$, and $(\cdot)^H$ is the Hermitian transpose of the vector. 
Here, it becomes apparent that the standard costs over states and inputs are constant for all frequencies. 
To leave the possibility of having different loopshaping per input dimension, the frequency-dependent weight matrix in \eqref{eq:FreqShapedCost} is generalized to
\begin{align*} 
    \tilde{\vR}(\omega) & = \! \begin{bmatrix}
    r^*_1(\omega) & & \\
    & \ddots & \\
    & & r^*_m(\omega)
  \end{bmatrix} \!  \vR \!   \begin{bmatrix}
    r_1(\omega) & & \\
    & \ddots & \\
    & & r_m(\omega)
  \end{bmatrix},
\end{align*}
\begin{align} 
r_i(\omega) & = \frac{1 + \beta_i j\omega}{1 + \alpha_i j\omega}, \qquad \beta_i>\alpha_i,
    \label{eq:individualweightingfunction}
\end{align}
where $r^*_i(\omega)$ is the complex conjugate of $r_i(\omega)$.
Every input direction can now have its own shaping function $r_i$.
In order to transfer this new cost function back into the time domain, a change of variables is required. 
The filtered inputs, $\hat{\vnu}(\omega)$, are defined elementwise as
\begin{align}
    \hat{\nu}_i(\omega) & = r_i(\omega) \hat{u}_i(\omega).
\end{align}
After substitution, we arrive back at a frequency-independent cost function over the filtered variables in \eqref{eq:J_filtered_frequencydomain}.
This cost is then converted back to the time domain in \eqref{eq:J_filtered_timedomain},
\begin{align}
    J &= \frac{1}{2\pi} \int_{-\infty}^{\infty} \left( \hat{\vx}(\omega)^H \vQ \, \hat{\vx}(\omega) 
    + \hat{\vu}(\omega)^H \tilde{\vR}(\omega)  \, \hat{\vu}(\omega) \right) d\omega
    \label{eq:J_filtered_frequencydomain_inbetween}  \\
    &= \frac{1}{2\pi} \int_{-\infty}^{\infty} \left( \hat{\vx}(\omega)^H \vQ \, \hat{\vx}(\omega) 
    + \hat{\vnu}(\omega)^H \vR \,\hat{\vnu}(\omega) \right) d\omega
    \label{eq:J_filtered_frequencydomain} \\
    &= \frac{1}{2} \int_0^{\infty} { \left( \vx(t)\!^T\! \vQ \, \vx(t) 
    + \vnu(t)\!^T \vR \, \vnu(t) \right)  dt}.
    \label{eq:J_filtered_timedomain}
\end{align}

\subsection{Implementation}
The presence of the filtered inputs $\vnu(t)$ in the cost function~\eqref{eq:J_filtered_timedomain} requires the augmentation of the original problem. 
Considering the original system dynamics, ${\dot{\vx} = \vf(\vx, \vu)}$, and state input constraint, $\vg(\vx, \vu) \leq \mathbf{0}$, this can be achieved in several ways. 
If $\vr(\omega)$ consists of proper rationals, its state space realization\footnotemark, \mbox{($\vA_r$, $\vB_r$, $\vC_r$, $\vD_r$)},  can be used to substitute for $\vnu$ in the cost function. The filter's internal dynamics are appended to the system.
\begin{equation}
    J = \frac{1}{2} \int_0^{\infty}  \left( \vx\!^T\! \vQ \, \vx   +  (\vC_{r} \vx_{r} + \vD_{r} \vu)\!^T\! \vR \, (\vC_{r} \vx_{r} + \vD_{r} \vu) \right)  dt 
\end{equation}
\begin{equation}
    \begin{bmatrix}
        \dot{\vx} \\
        \dot{\vx}_r
    \end{bmatrix} =
        \begin{bmatrix}
        \vf(\vx, \vu)  \\
        \vA_r  \vx_{r} + \vB_r \vu
    \end{bmatrix},  \quad  \vg(\vx, \vu) \leq \mathbf{0},
\end{equation}
If $\vr(\omega)$ consists of improper rationals, the transfer function $\vs(\omega) = \vr^{-1}(\omega)$ is defined such that $\hat{u}_i(\omega) = s_i(\omega) \hat{\nu}_i(\omega)$. The state space realization\footnotemark[\value{footnote}] of $\vs(\omega)$, ($\vA_s$, $\vB_s$, $\vC_s$, $\vD_s$),  is used to substitute for $\vu$ in both system dynamics and constraints. Again, the filter's internal dynamics are added to the system.
\footnotetext{State space realizations are computed according to the balanced realization described in \cite{Moore1981}}
\begin{equation}
J = \frac{1}{2} \int_0^{\infty} { \left( \vx\!^T\! \vQ \, \vx 
+ \vnu\!^T \vR \, \vnu \right)  dt} 
\end{equation}
\begin{equation}
\begin{bmatrix}
\dot{\vx}  \\
\dot{\vx}_s
\end{bmatrix} =
\begin{bmatrix}
\vf(\vx, \vC_{s} \vx_{s} \!+\! \vD_{s} \vnu)  \\
 \vA_s  \vx_{s} + \vB_s \vnu
\end{bmatrix},  \,\, \vg(\vx, \vC_{s} \vx_{s} \!+\! \vD_{s} \vnu) \leq \mathbf{0}, \label{eq:loopshaping_dynamics_constraints} 
\end{equation}
The system inputs, $\vu = \vC_{s} \vx_{s} + \vD_{s} \vnu$, are then retrieved after optimization.
Because the selected class of shaping functions are of relative degree zero, choosing between these equivalent options is a numerical consideration. 
Since the poles of each $r_i(\omega)$ are in our case higher than its zeros, the dynamics of $\vA_r$ are faster than those of the realization of its inverse, $\vA_s$, we therefore use the latter formulation.

\subsection{Related methods}
\label{sect:relatedmethods}
\begin{figure}[tb]
    \centering
    \includegraphics[width=1.0\columnwidth]{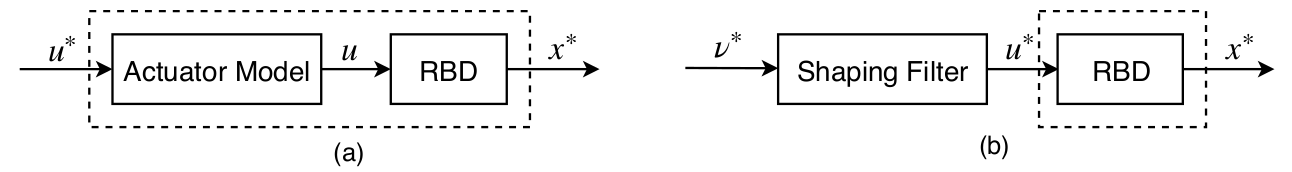}
    \vspace{-7mm}
    \caption{Block diagrams for actuator modeling (a) and frequency shaping (b). RBD denotes the rigid body dynamics. $\vu^*$ and $\vx^*$ are the optimized system input and state references to be tracked by the robot.}
    \label{fig:systemmodeling}
    \vspace{-5mm}
\end{figure}

The relation with higher order formulations can be understood by considering a shaping function of $r_i(\omega) = j\omega$ for each input in~\eqref{eq:individualweightingfunction} instead of $\frac{1+\beta_i j\omega}{1+\alpha_i j\omega}$.
The resulting state space realization is
\begin{equation}
\dot{\vx}_s = \vnu, \qquad \vu = \vx_s, \\
\end{equation}
which is equivalent to $\dot{\vu} = \vnu$, \ie the auxiliary input is the derivative of the original input.
By selecting $r_i(\omega) = (j\omega)^n$, general n-th order formulations can be retrieved.
Higher order methods can thus be seen as a special case of frequency shaping. 
The proposed method considers more general transfer functions, allowing the user with more flexibility to target a specific frequency range.
There is, however, a numerical consideration when using frequency shaping functions that go to infinity as $\omega$ goes to infinity. 
For such an transfer function $\vs(\omega)$ will be strictly proper and have a realization with $\vD_s = 0$. 
As can be seen from the constraints in \eqref{eq:loopshaping_dynamics_constraints}, this results in pure-state constraints. 
This is not a problem in theory, but such constraints are computationally more expensive to handle than state-inputs constraints, so we avoid them in practice to achieve fast replanning.
When using a higher order formulation, inequality constraints can be used to put hard limits on the smoothness of the trajectories.
Similar constraints can be placed in the frequency based method as done in \cite{hours2015constrained}.
Such a discussion is thus rather a preference for designing behavior through costs or constraints.
The difference between the proposed method and embedding an actuator model can be understood from Fig.~\ref{fig:systemmodeling}.
When embedding an actuator model in the system dynamics, one would interpret the input to that model as the command to be sent to the robot.
However, in the proposed method, the command sent to the robot is the original input, leaving the assumed relation between $\vx$ and $\vu$ unchanged.
In the former, the optimized state input trajectories, $\{\vx^*, \vu^*\}$ relies on the accuracy of the actuator model, while in the latter, the filter is used to restrict input frequency content to a feasible range.

\section{Experimental setup}

\subsection{Problem Formulation} 
The proposed method is applied to the kinodynamic model of a quadruped robot, which describes the dynamics of a single free-floating body along with the kinematics for each leg. The Equations of Motion (EoM) are given by
\begin{align*}
&\left\{ 
\begin{array}{ll}
		\dot{\vth} = \vT(\vth) \vom \\
		\dot{\vp}  = _W\!\vR_B(\vth) \, \vv \\
		\dot{\vom} = \vI^{-1} \left(  -\vom \times \vI \vom + \sum_{i=1}^4{ {\vr_{EE}}_i(\vq) \times {\vlambda_{EE}}_i} \right)\\
		\dot{\vv}  = \vg(\vth) + \frac{1}{m} \sum_{i=1}^4{\vlambda_{EE}}_i \\
		\dot{\vq} = \vu_J,
\end{array}	 
\right. \notag
\end{align*}
where $_W\!\vR_B$ and $\vT$ are the rotation matrix of the base with respect the global frame and 
the transformation matrix from angular velocities in the base frame to the Euler angles derivatives in the global frame.
$\vg$ is the gravitational acceleration in body frame, $\vI$ and $m$ are the moment of inertia about the CoM and the total mass respectively. The inertia is assumed to be constant and taken at the default configuration of the robot. ${\vr_{EE}}_i$ is the position of the foot $i$ with respect to CoM. 
$\vth$ is the orientation of the base in Euler angles, $\vp$ is the position of the CoM in world frame, $\vom$ is the angular rate, and $\vv$ is the linear velocity of the CoM. $\vq$~is the vector of twelve joint positions. The inputs of the model are the joint velocity commands $\vu_J$ and end effector contact forces ${\vlambda_{EE}}_i$. 

The constraints depending on the mode of a leg at that point in time are formulated as
\begin{align*}
&\left\{ 
\begin{array}{ll}
		{\vv_{EE}}_i = \mathbf{0}, \quad  {\vlambda_{EE}}_i \in \mathcal{C}(\hat{\vn}, \mu), \quad &\text{if $i$ is a stance leg} \\
		{\vv_{EE}}_i \cdot \hat{\vn} = c(t), \quad  {\vlambda_{EE}}_i = \mathbf{0}, \quad  &\text{if $i$ is a swing leg} 
\end{array}
\right.
\end{align*}
where ${\vv_{EE}}_i$ is the end effector velocity in world frame, which constrains a stance leg to remain on the ground and a swing leg to follow the predefined curve $c(t)$ in the direction of the local surface normal, $\hat{\vn}$, to avoid foot scuffing with zero contact force inputs ${\vlambda_{EE}}_i$. This curve ends with a negative velocity of \SI{0.75}{\meter/\second}, which is maintained until contact is detected. The friction cone, $\mathcal{C}(\hat{\vn}, \mu)$, is defined by the surface normal and friction coefficient, $\mu = 0.7$. This constraint is enforced by projecting the inputs onto the feasible set in the forward rollout of the SLQ-MPC algorithm. Limitations of such a clamping strategy are discussed in \cite{tassa2014control}. In practice, we find that these constraints are rarely active and that the projection is sufficient for the motions in this work.

The baseline cost is formulated as a quadratic function
\begin{align}
J =&  \int_{0}^{T} L(\vx(t),\vu(t)) \, dt + \Phi(\vx(T)) \notag \\
L =& \frac{1}{2} \left( \vx-\vx_d \right)\!^T\! \vQ \left( \vx-\vx_d \right)  + \frac{1}{2} (\vu-\vu_0)\!^T \vR (\vu-\vu_0) \notag \\
\Phi =&  \frac{1}{2}\left( \vx-\vx_d \right)\!^T\! \vQ_T \left( \vx-\vx_d \right) ,
 \label{eq:mpc_cost}
\end{align}
where $\vx_d = [\vth_{des}^T, \vp_{des}^T, \vom_{des}^T, \vv_{des}^T, \vq_0^T]^T$ is a desired state consisting of commanded base pose and twist by the user and a default configuration for the joints.
Inputs are defined as $\vu = [{\vlambda_{EE}}_1^T, \dots, {\vlambda_{EE}}_4^T, \vu_j^T]$, and
$\vu_0$ is the equilibrium input for standing in the default configuration.
$\Phi(\cdot)$ is the final state cost, which is a heuristic to approximate the truncated infinite horizon, and is implemented as a diagonal cost on the base pose and velocities.
$L(\cdot,\cdot)$ is the intermediate cost where we use a simple diagonal cost on all state variables and contact force inputs.
For the costs on the joint velocities, a diagonal matrix is pre- and post-multiplied by the end-effector Jacobians to define costs over the task space.

\subsection{System integration}
In the model described in the previous section, the control inputs are end-effector forces and joint velocities.
To translate the solution to torque commands, we extract a full position, velocity, and acceleration plan for CoM and end-effector trajectories, in addition to the planned contact forces.
This plan is tracked by the hierarchical whole-body control (WBC) architecture described in \cite{bellicoso2016perception}.
The tasks in decreasing priority are (1) satisfying the equation of motions and zero acceleration for contact feet (2) tracking CoM and swing leg trajectories, and (3) tracking the planned contact forces. 
The desired contact forces from the MPC are thus communicated in two ways:
The CoM trajectory dictates the net acting forces, and force tracking task regulates the internal forces.
Without the latter, contact forces would be redistributed among the contacts, which would override the planned smoothness of the trajectories.
Additionally, on all priorities, torque limits and friction cone constraints are imposed as inequality constraints.

\begin{figure}[tb]
    \centering
    \includegraphics[width=1.0\columnwidth]{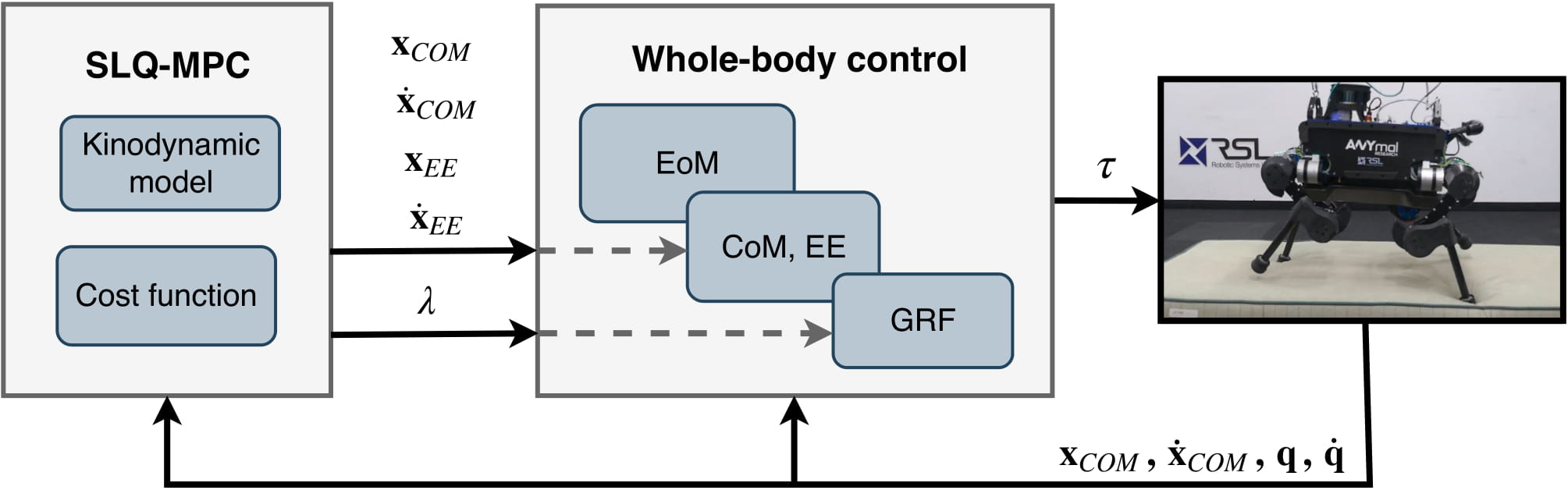}
    \caption{MPC and whole-body control structure overview. The SLQ-MPC algorithm running on a separate desktop PC sends center of mass (CoM) and end-effector (EE) reference to the onboard whole-body control structure. This hierarchical controller computes torque commands based on the listed priorities.}
    \label{fig:systemIntegration}
    \vspace{-4mm}
\end{figure}

The optimal control problem in \eqref{eq:mpc_cost} is solved for a user-defined gait with the continuous time SLQ-MPC algorithm described in \cite{farshidian2017efficient}.
We use a receding horizon length of $1.0$s, which results in an MPC update rate of $70$Hz for the baseline method and around $40$Hz for the frequency-shaped method.
The MPC runs on a desktop PC with an Intel Core i7-8700K CPU@3.70GHz hexacore processor and continuously computes a motion plan from the latest known state through a real-time-iteration scheme. 
The WBC runs on the dedicated onboard PC and tracks the most recent plan.
Here, the augmented filter state is propagated as well with the currently available augmented input plan $\vnu(t)$.
Both nodes communicate over a local network.
An overview of the experimental setup is provided in Fig.~\ref{fig:systemIntegration}.

\section{Results}
We study the effects of adopting the cost function in \eqref{eq:J_filtered_timedomain} for the previously described setup under various locomotion tasks.
To see the results at different levels of model errors, we conduct perfect model simulations, rigid-body simulations, and hardware experiments.
When selecting different values for $\beta$, $\alpha$ is selected such that the frequencies in the stopband incur a cost of $100$ times the steady state cost, \ie $\alpha = 0.1\beta$.

\subsection{Perfect model}
\label{sect:planning}
\begin{figure}[tb]
    \centering
    \includegraphics[width=0.7\columnwidth]{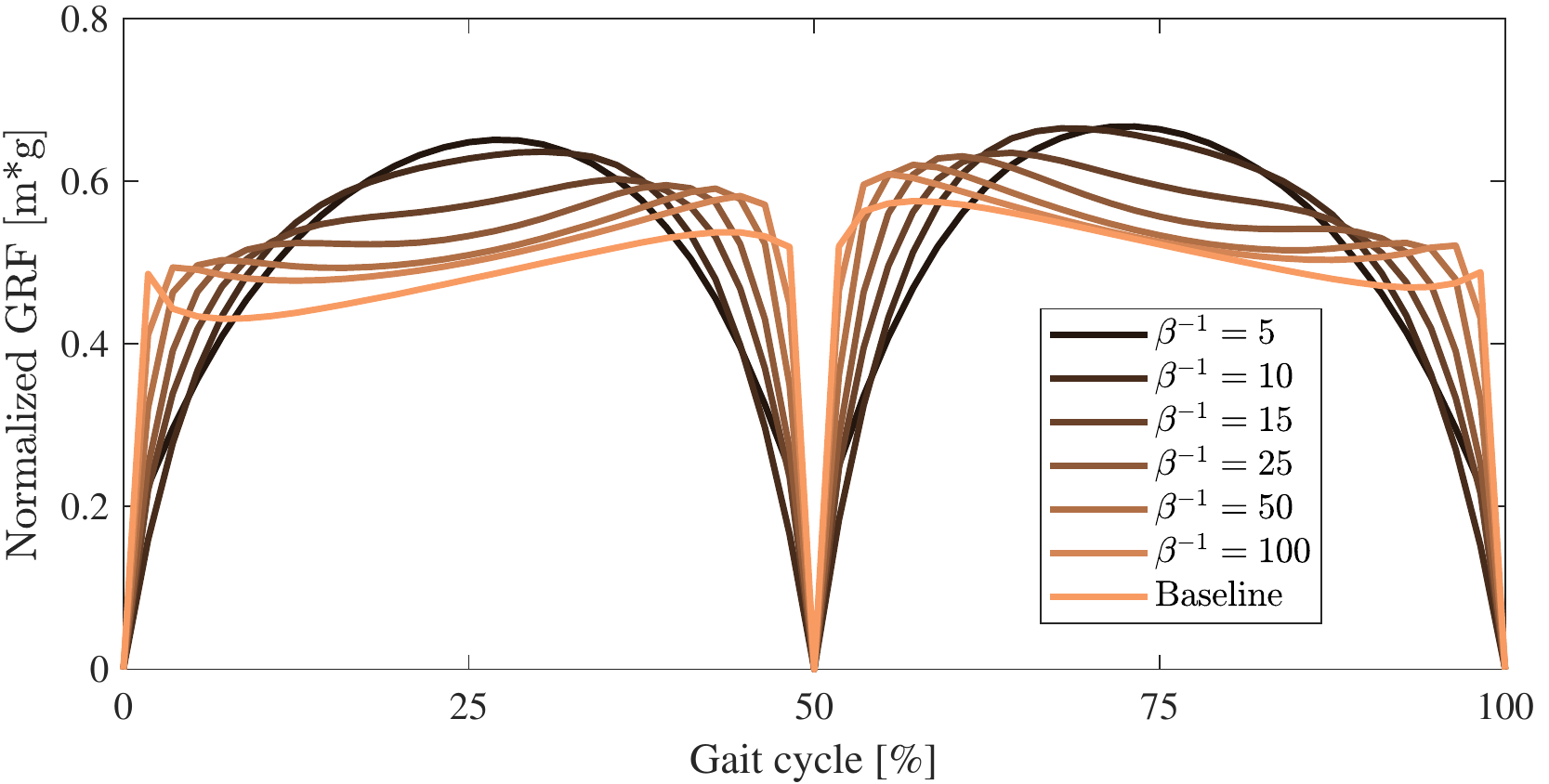}
    \vspace{-3mm}
    \caption{Ground reaction forces in z-direction for a trotting gait at \SI{0.5}{\meter/\second} with a period of $\SI{0.7}{\second}$. The first half of the gait cycle corresponds to the left front foot and the second half to the left hind foot. As the frequency limit decreases from infinity (\ie baseline) to 5, the smoothness of the planned contact forces increases.}
    \label{fig:gaitGRFBeta}
\end{figure}

\begin{figure}[tb]
    \centering
    \includegraphics[width=0.7\columnwidth]{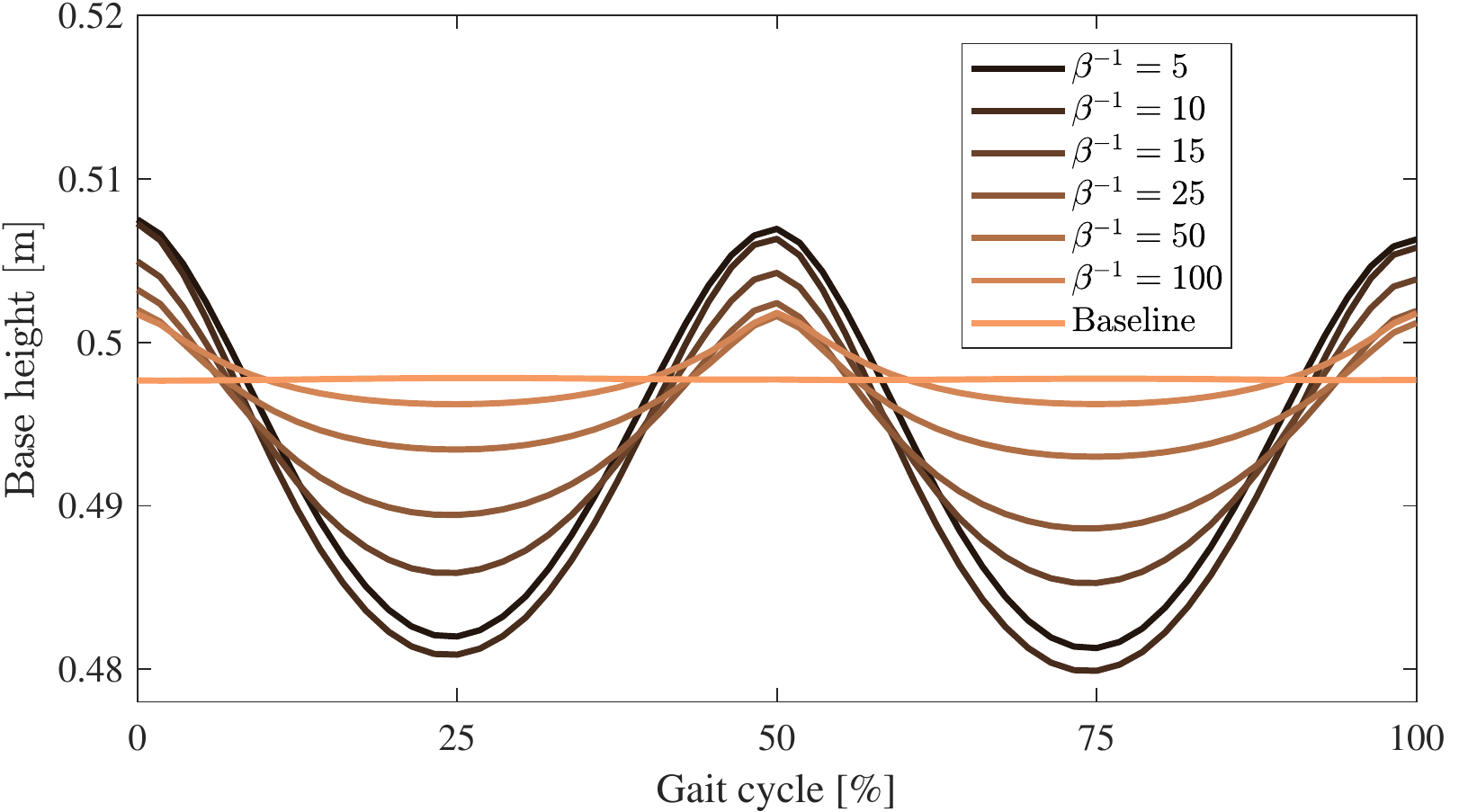}
    \vspace{-3mm}
    \caption{Base height for a trotting gait at \SI{0.5}{\meter/\second} with a period of $\SI{0.7}{\second}$ under different smoothing parameters. In general, the vertical displacement of the base increases as the controller becomes smoother since it cannot abruptly increase or decrease the ground reaction forces at the stance feet.}
    \label{fig:gaitBaseZBeta}
    \vspace{-4mm}
\end{figure}

First, we investigate the effect of the loopshaping on the contact forces in a simulation that uses the same model as the MPC.
This shows how the resulting trajectories are different already in the case of no modeling errors.
The analyzed gait is a trot with a duty factor of 0.5, \ie with no overlap in stance phases of the diagonal feet, while a  forward velocity of \SI{0.5}{\meter/\second} is commanded.
Fig.~\ref{fig:gaitGRFBeta} shows the ground reaction forces when selecting different values for $\beta$. 
As seen from the plot, the baseline method instantaneously applies contact forces once a foot is in contact.
As expected, lowering the frequency at which costs start to increase, \ie lowering $\beta^{-1}$, results in increasingly smooth trajectories.
The frequency-shaped method approaches the baseline as $\beta^{-1}$ goes to infinity.
The corresponding base height trajectories are plotted in Fig.~\ref{fig:gaitBaseZBeta}.
Smoother contact force trajectories require more vertical displacement of the base, while the baseline produces the exact amount of force to keep the base level.

\subsection{Physics simulation}
\begin{figure}[tb]
    \centering
    \includegraphics[width=0.95\columnwidth]{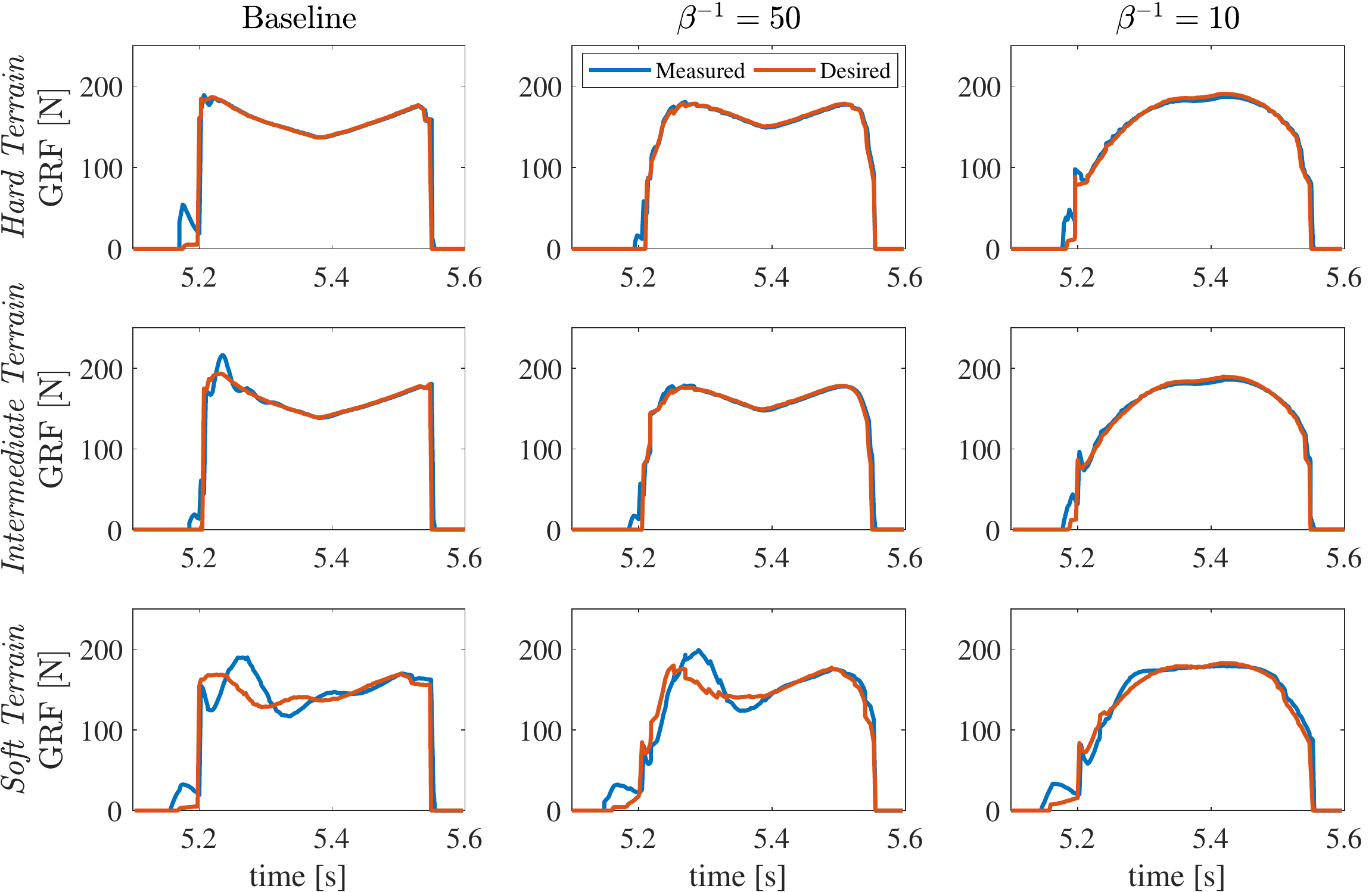}
    \vspace{-3mm}
    \caption{Measured and desired contact forces in z-direction during a trot in-place for various ground properties and cost functions. The columns from left to right correspond to the baseline, $\beta^{-1}=50$, and $\beta^{-1}=10$ cost functions. The rows, from top to bottom corresponds to ground properties with $k_p$ and $k_d$ of $\{1e6, 100\}$, $\{1e5, 50\}$, and $\{1e4, 30\}$. As the compliance of the terrain increases, the difference between desired forces generated from the whole-body controller and resulting forces grows.
    }
    \label{fig:contactForces}
    \vspace{-4mm}
\end{figure}

The combination of tracking controller and MPC is evaluated in the Open Dynamics Engine (ODE) \cite{smith2005open} rigid-body simulation, 
where we can vary ground properties in a controlled way.
The model errors, in this case, come from the difference between rigid-body dynamics and the kinodynamic model, as well as the assumption of a rigid ground contact when the terrain is made compliant.
ODE allows for simulation of soft contacts by modeling\footnote{ODE relaxes the rigid-contact solver such that it implicitly resembles a spring-damper interaction.} the ground contact forces as a spring-damper system.
Three different sets of spring-damper parameters $k_p$ and $k_d$ are selected to simulate hard, intermediate, and soft ground, respectively.
For each terrain, three different cost functions are evaluated: the baseline without frequency shaping as well as frequency-dependent cost functions with $\beta^{-1}=50$ and $\beta^{-1}=10$.
These values were selected based on Fig.~\ref{fig:gaitGRFBeta} to represent three levels of smoothness in the continuum of available cost functions.
The resulting contact force profiles for all combinations are shown for a single stance phase in Fig.~\ref{fig:contactForces}.
Desired and measured contact forces are shown for a single leg during an in-place trotting motion with a stance duration of $0.35$ seconds.
As the compliance of the terrain increases, the difference between desired forces generated from the WBC and resulting forces grows.
The WBC uses rigid-body dynamics with a hard contact assumption to compute desired contact forces.
The difference between desired and measured forces is therefore a measure of disturbances inserted by additional unmodeled dynamics, which in general includes the bandwidth limits of actuators and contact dynamics that we aim to avoid.
Table~\ref{tab:contactForces} shows the force tracking performance averaged over six gait cycles in Mean Absolute Error (MAE) and Mean Squared Error (MSE) defined as
\begin{equation}
\text{MAE} = \frac{1}{T} \int_{0}^{T} | \vlambda - \vlambda_{des} | dt, \quad  \text{MSE} = \frac{1}{T} \int_{0}^{T} (\vlambda - \vlambda_{des})^2 dt. \notag
\end{equation}
For all cost functions, tracking performance degrades as the model error increases. 
Qualitatively, the baseline controller suffers from a larger error at the beginning and end of the contact phase, due to its step-like change in the desired forces.
Even on hard terrain, there is an apparent benefit of loopshaping.
The smoother transition between contact phases mitigates the disturbance generated by contact timing mismatch.
Differences become most apparent for the soft terrain, where the smoother trajectory has better performance especially in MSE. 

\begin{table}[bt]
\centering
\caption{Force tracking performance MAE (MSE) [N~(N$^2$)] for different terrain and cost functions in simulation.}
\begin{tabular}{llll}
\toprule
& \multicolumn{3}{c}{Cost function}               \\ \cmidrule{2-4} 
Terrain $\{k_p, k_d\}$     & Baseline     & $\beta^{-1}=50$ & $\beta^{-1}=10$ \\ \midrule
Hard $\{1e6, 100\}$  & 4.8 (303.5) & \textbf{3.6} (58.5)     & 5.0 (104.2)     \\ 
Medium $\{1e5, 50\}$ & 5.5 (316.2)  & \textbf{4.6} (110.1)     & 5.0 (100.4)     \\ 
Soft $\{1e4, 30\}$  & 13.5 (525.5) & 11.6 (286.9)    & \textbf{7.4} (146.5)  \\ \bottomrule
\end{tabular}
\label{tab:contactForces}
\vspace{-3mm}
\end{table}

\begin{figure}
\centering
  \begin{minipage}[t]{\columnwidth}
  \centering
    \includegraphics[width = 0.9\textwidth]{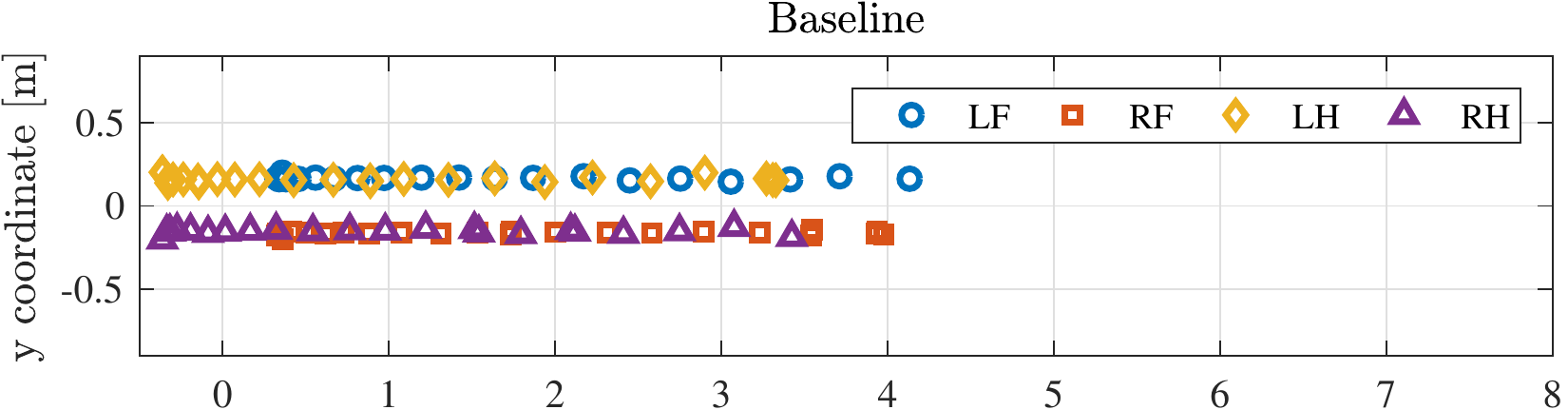}
  \end{minipage}
  \hfill
  \begin{minipage}[t]{\columnwidth}
  \centering
    \includegraphics[width = 0.9\textwidth]{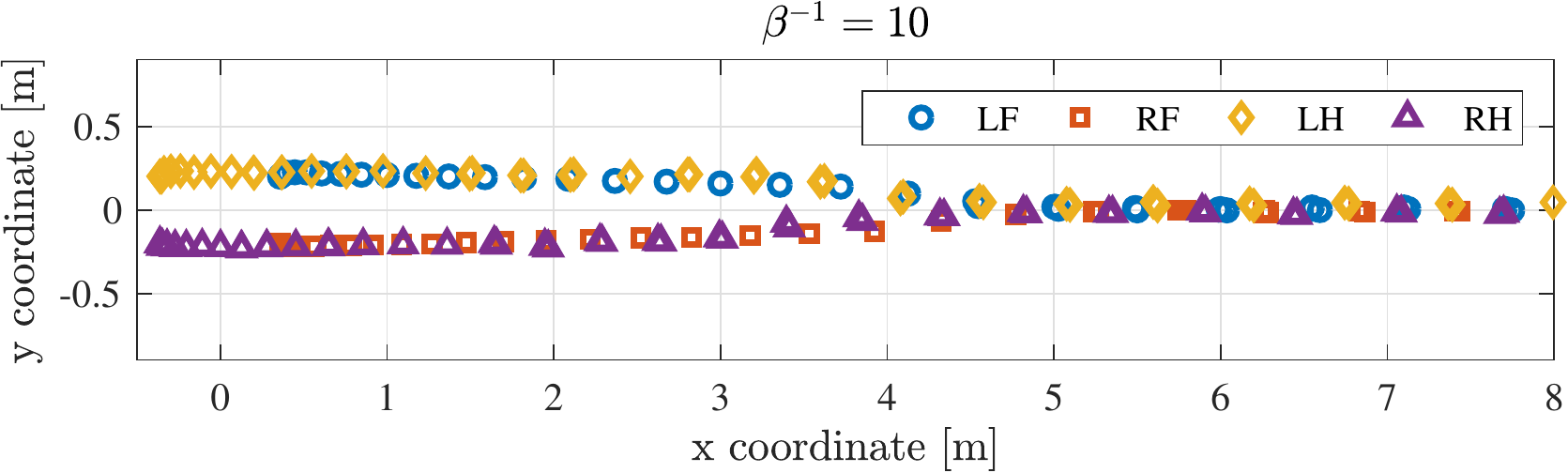}
  \end{minipage}
  \vspace{-5mm}
  \caption{Foot placement strategy for the baseline (top) and the $\beta^{-1}=10$ (bottom) cost functions during a trot with increasing velocity commands in the positive x-direction. Commands start at $v=\SI{0}{\meter/\second}$ and accelerate with \SI{0.05}{\meter/\second\squared}. Failure occurs at \SI{0.6}{\meter/\second} and \SI{0.9}{\meter/\second} respectively. The footstep strategy of the baseline method is different from the frequency-aware solution. While the former minimizes the lateral motion of the center of pressure by aligning the support polygons (lines), the baseline method does not adapt its footstep planning based on the velocity.}
  \label{fig:footPlacement}
\end{figure}

Furthermore, we examine the locomotion strategy under two extrema of cost functions.
The commanded forward velocity during a trotting motion is gradually increased until failure occurs.
The foot placement strategies are visualized in Fig.~\ref{fig:footPlacement}. 
The plots show footstep locations from a top-view with the robot starting at the origin.
As the velocity increases towards the right side of the plot, the footstep locations start to differ.
Interestingly, with the smoother cost function of $\beta^{-1}=10$, the foot placement strategy is significantly altered and becomes velocity dependent.
Where under the baseline costs the controller chooses a fixed width foot placement, the frequency-shaped solution places the feet increasingly inwards for higher velocities.
This can be explained by realizing that horizontal forces, like the normal forces in Fig.~\ref{fig:gaitGRFBeta}, smoothly start from and end at zero. 
During a switch from one contact pair to the next, a lateral force is required to make the CoM velocity change in the direction of the next support line.
Under the frequency-dependent cost function, high lateral forces around the contact switch are expensive, and a solution where the supports are more aligned is thus preferred.
As the forward velocity increases, so does the required change in lateral velocity for a given width. The alignment of support lines therefore gradually becomes more pronounced.

Remarkably, this results in a significantly higher maximum velocity of \SI{0.9}{\meter/\second} versus \SI{0.6}{\meter/\second}.

\subsection{Hardware}
\begin{figure*}[tb]
    \centering
    \includegraphics[clip,trim={0mm 0mm 0mm 0mm},width=.95\textwidth]{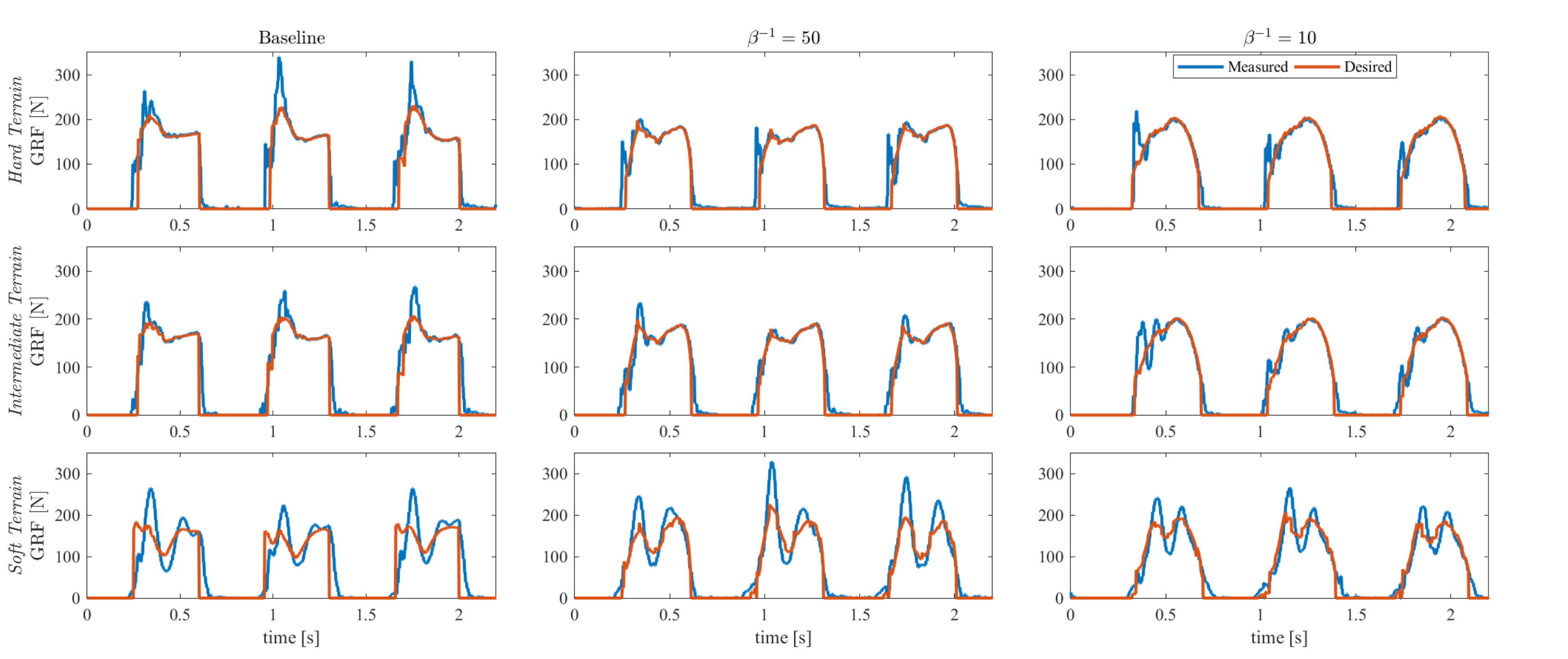}
     \vspace{-4mm}
    \caption{Measured and desired contact forces for the right front foot during the first three trotting gait cycles on the lab floor (top row), foam (middle row) and a mattress (bottom row) with the baseline (left column), $\beta^{-1}=50$ (middle column), and $\beta^{-1}=10$ (right column) cost functions. A smoother transition between swing and stance phase results in better tracking performance.}
    \label{fig:contactForcesHardware}
    \vspace{-4mm}
\end{figure*}

\begin{figure}[tb]
    \centering
    \includegraphics[clip,trim={0mm 0mm 0mm 0mm},width=1.0\columnwidth]{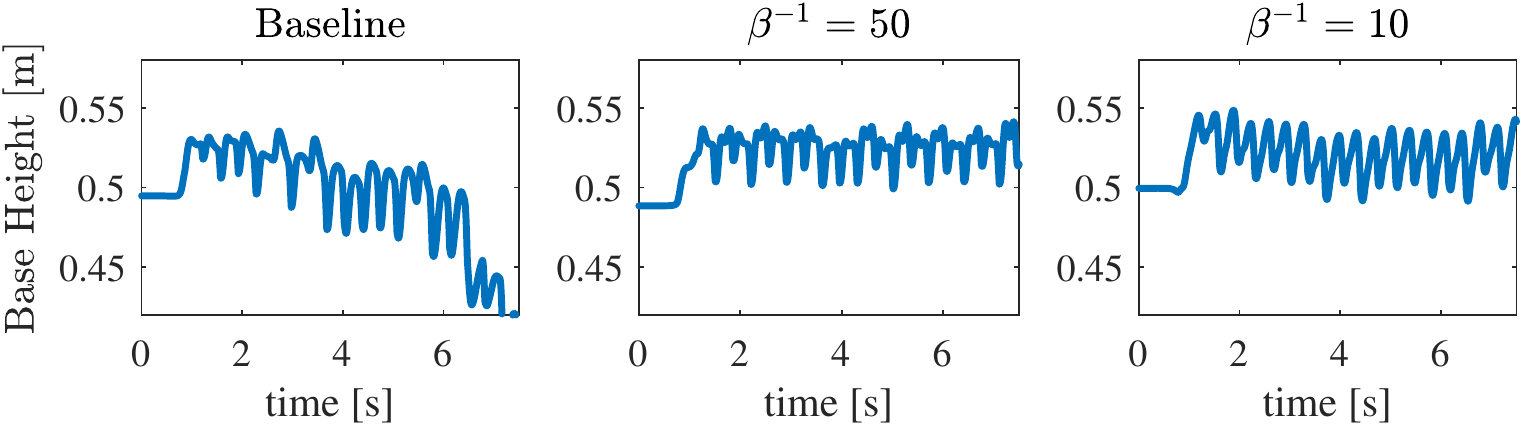}
    \vspace{-6mm}
    \caption{Measured base height during the first seconds of trotting motion on the mattress with the baseline (left column), $\beta^{-1}=50$ (middle column), and $\beta^{-1}=10$ (right column) cost functions. Where the baseline method fails to maintain its height, the frequency-aware controllers remain stable.}
    \label{fig:baseHeightHardware}
    \vspace{-5mm}
\end{figure}

On hardware, we aim to validate the simulation results for contact force tracking performance on different terrains;
The floor of the lab, a \SI{3.5}{\centi\meter} foam block, and a mattress are selected to test a rigid, an intermediate and a very compliant terrain respectively.
A force-torque sensor is mounted on the right front leg to obtain direct measurements of the ground reaction forces.
The resulting measured and desired forces for different cost function and terrain combinations are shown in Fig.~\ref{fig:contactForcesHardware}.
The plots show the difference between measured and desired contact forces of the right-front leg during the first three steps.
The MAE and MSE averaged over those first three gait cycles are given in Table~\ref{tab:contactForcesHardware}.
On hard terrain, all methods perform well, and slight differences in tracking performance occur at the beginning and end of the contact phase.
In this area the $\beta^{-1}=10$ controller provides the smoothest transition, resulting in the best MAE and MSE on all terrains.
Where in the simulation experiments we see that a medium amount of smoothing is best for hard and medium stiff terrain, we do not see this in the hardware experiments.
The difference could be caused by the series elastic actuators of ANYmal, causing model errors and bandwidth limits even on hard terrain.
For a step input to the torque level reached during trot, a $90\%$ rise-time up to \SI{35}{\milli\second}, equivalent to a bandwidth of around \SI{60}{\radian/\second} can be expected \cite{hutter2016anymal}.
When further reducing the compliance by trotting on the mattress, the baseline and $\beta^{-1}=50$ controllers suffer a substantial decrease in performance.
The base height during the first part of the experiment is shown in Fig.~\ref{fig:baseHeightHardware}.
Due to the significant mismatch between the planned and resulting contact force with each footstep, the baseline controller loses base height in a few steps, causing it to fail.
The $\beta^{-1}=50$ cost function does achieve a trot, as shown in the video\footnote{\url{https://youtu.be/RSJgqkk2VRI}}, but strong oscillations are present between the terrain and the feet.
The $\beta^{-1}=10$ cost function, finally, results in both a stable trot and a smooth contact interaction.

\begin{table}[bt]
\centering
\caption{Force tracking performance MAE (MSE) [N~(N$^2$)] for different terrain and cost functions on hardware.}
\begin{tabular}{llll}
\toprule
& \multicolumn{3}{c}{Cost function}               \\ \cmidrule{2-4} 
Terrain    & Baseline     & $\beta^{-1}=50$ & $\beta^{-1}=10$ \\ \midrule 
Hard   & 27.1 (1465.7) & 14.7 (785.5) & \textbf{12.7} (641.4)   \\
Medium & 21.4 (1040.1) & 19.7 (741.3) & \textbf{16.0} (555.4)\hphantom{0}   \\
Soft   & 30.9 (2308.1) & 26.0 (1237.5) & \textbf{22.0} (824.3)   \\ \bottomrule
\end{tabular}
\label{tab:contactForcesHardware}
\vspace{-4mm}
\end{table}

In the accompanying video, we additionally show the behavior under disturbances.
The robot trots in place and has costs on base deviation from the initial position.
We disturb the robot in the horizontal plane.
Qualitatively, the reactive stepping and push-back behavior differ.
Under the baseline method, the robot reacts to a push by generating lateral forces and the user experiences instant resistance.
The frequency-aware implementation with $\beta^{-1}=10$ instead accepts deviation of the base trajectory and adapts future step location and force profile to smoothly return to the origin.

\section{Discussion}
We have shown that a single parameter of a frequency-dependent cost function provides a handle on a rich variety of solutions.
While the smoothness at the beginning of the stance phase is not surprising due to the filter, the anticipatory decrease in force before lifting the foot, as seen in Fig.~\ref{fig:gaitGRFBeta}, shows that the filter and planning are tightly working together.
The additional filter states allow the Riccati-type algorithm to reason about future state-input constraints, in this case, zero contact forces during the swing phase, and adapts the strategy to approach them smoothly.
This is remarkable because the backward pass is projected on those constraints only at the point in time that they are active.
Interestingly we also find that the foot placement strategy changes significantly.
These observations highlight the fact that embedding frequency awareness of the MPC is richer than simply filtering the obtained inputs. 
As a future work, we will explore ways to change to cost function online and in this way adapt the locomotion strategy to the terrain.
For the legs in swing phase, optimizing over joint torque, which is effectively one derivative higher than optimizing over joint velocities, can also induce a smoother swing trajectory. 
However, obtaining real time performance with the extra nonlinear effects is challenging. We believe that keeping the model as simple as possible helps to get a robust solution working on hardware.

\section{Conclusion}
We introduced frequency-aware MPC by combining frequency-dependent cost functions with modern MPC methods.
With simulation experiments, we show that the resulting smoother force profiles improve tracking performance when the rigid terrain assumption is relaxed, without the need to explicitly model it.
We validated these results on hardware and see a similar trend when comparing performance on various terrains.
The method is shown to provide robustness against unmodeled dynamics of series elastic actuators and compliant terrains.
We demonstrated that with this approach ANYmal is now able to execute dynamic motions even on highly compliant terrains.

\bibliographystyle{bibtex/myIEEEtran} 
\bibliography{bibtex/IEEEabrv,bibtex/IEEEexample,bibtex/bibliography}

\end{document}